\documentclass[letterpaper, 10 pt, conference]{ieeeconf}  

\IEEEoverridecommandlockouts                              

\usepackage{graphics} 
\usepackage{graphicx}

\usepackage{epsfig} 
\usepackage{amsmath} 
\usepackage{amssymb}  

\usepackage[binary-units]{siunitx}

\usepackage{algorithm}
\usepackage[noend]{algpseudocode} 
\algnewcommand\AAND{\textbf{ and }}
\algnewcommand\Or{\textbf{ or }}

\usepackage{color}
\usepackage{citesort}
\usepackage{url}
\usepackage{breakurl}
\usepackage[breaklinks]{hyperref}

\DeclareMathAlphabet{\pazocal}{OMS}{zplm}{m}{n}

\newcommand{\Ys}{\pazocal{Y}}

\def \*#1 {mathbf{#1}}
\def \@#1 {\mathbb{#1}}

\newtheorem{definition}{Definition}

\DeclareMathAlphabet{\mathpzc}{OT1}{pzc}{m}{it}

\usepackage{array}
\newcolumntype{C}[1]{>{\centering\arraybackslash}p{#1}}
\newcolumntype{M}[1]{>{\raggedright\arraybackslash}p{#1}}

\usepackage{array} 
\newcolumntype{L}[1]{>{\raggedright\let\newline\\\arraybackslash\hspace{0pt}}m{#1}}	
\newcolumntype{S}[1]{>{\centering\let\newline\\\arraybackslash\hspace{0pt}}m{#1}}
\newcolumntype{R}[1]{>{\raggedleft\let\newline\\\arraybackslash\hspace{0pt}}m{#1}}

\newtheorem{problem}{Problem}

\usepackage[nolist,nohyperlinks]{acronym}
\acrodef{fov}[FoV]{Field of View}
\acrodef{tsp}[TSP]{Traveling Salesman Problem}
\acrodef{gvi}[GVI]{General Visual Inspection}
\acrodef{ve}[VE]{Volumetric Exploration}
\acrodef{esdf}[ESDF]{Euclidean Signed Distance Field}
\acrodef{fpso}[FPSO]{Floating Production Storage and Offloading}
\acrodef{iacs}[IACS]{International Association of Classification Societies}

\makeatletter
\renewcommand*{\@opargbegintheorem}[3]{\trivlist
  \item[\hskip \labelsep{\itshape #1\ #2}] \textit{(#3)}\ }
\makeatother

\title{\LARGE \bf
Autonomous Exploration and General Visual Inspection of Ship Ballast Water Tanks using Aerial Robots
}

\author{Mihir Dharmadhikari$^{1}$, Paolo De Petris$^{1}$, Mihir Kulkarni$^{1}$, Nikhil Khedekar$^{1}$, Huan Nguyen$^{1}$,\\ 
Arnt Erik Stene$^{2}$, Eivind Sj{\o}vold$^{2}$,
Kristian Solheim$^{3}$, Bente Gussiaas$^{3}$, and Kostas Alexis$^{1}$
\thanks{This material was supported by the Research Council of Norway under project SENTIENT (NO-321435).}
\thanks{$^{1}$The authors are with the Autonomous Robots Lab, Norwegian University of Science and Technology (NTNU), O. S. Bragstads Plass 2D, 7034, Trondheim, Norway {\tt\small mihir.dharmadhikari@ntnu.no}}
\thanks{$^{2}$The authors are with Equinor ASA, Norway {\tt\small arnts@equinor.com}}
\thanks{$^{3}$The authors are with Altera Infrastructure Production AS, Norway {\tt\small bente.gussiaas@alterainfra.com}}
}

\begin{document}

\maketitle
\thispagestyle{empty}
\pagestyle{empty}

\begin{abstract}
This paper presents a solution for the autonomous exploration and inspection of Ballast Water Tanks (BWTs) of marine vessels using aerial robots. Ballast tank compartments are critical for a vessel's safety and correspond to confined environments often connected through particularly narrow manholes. The method enables their volumetric exploration combined with visual inspection subject to constraints regarding the viewing distance from a surface. We present evaluation studies in simulation, in a mission consisting of 18 BWT compartments, and in 3 field experiments inside real vessels. The data from one of the experiments is also post-processed to generate semantically-segmented meshes of inspection-important geometries. Geometric models can be associated with onboard camera images for detailed and intuitive analysis. 
\end{abstract}

\section{INTRODUCTION}\label{sec:intro}
Ships and other marine structures, such as maritime oil and gas infrastructure, can endure harsh conditions over long lifetimes. Corrosion and structural degradation have to be carefully monitored to ensure integrity. With the world greatly relying on maritime transport and marine resources, a global fleet of approximately $54,000$ large ($>1,000$ gross tons ~\cite{sirimanne2019review}) maritime structures are mainly inspected manually by human surveyors,  while the broader global fleet involves more than $100,000$ ships~\cite{handbookstatistics2022}. 
Among others, the surveyors must inspect the Ballast Water Tanks (BWTs) which represent dangerous, confined, enclosed environments often with difficult access via narrow hatches and manholes, low-lighting, slippery surfaces, as well as possible oxygen deficiency or presence of toxic gases. 
The European Maritime Safety Agency (EMSA) reports that a significant number of accidents aboard ships between 2014-2021 were due to the fall of persons (e.g., within the challenging enclosed ballast tank and cargo hold facilities)~\cite{emsa2022annual}. Furthermore, due to the complexities and laborious nature of ship inspection processes, they often take place at select ``dry docks'' which in turn implies that ships often spend multiple weeks out of operation either because they have to sail to specific dry docks or because supporting activities such as scaffolding preparations must take place before an inspection commences. This leads to great financial losses (with just one day on dry dock costing approximately $\$1$ million~\cite{bonnin2016generic}) and loss of opportunity (due to dead time). 

\begin{figure}[ht!]
\centering
    \includegraphics[width=0.99\columnwidth]{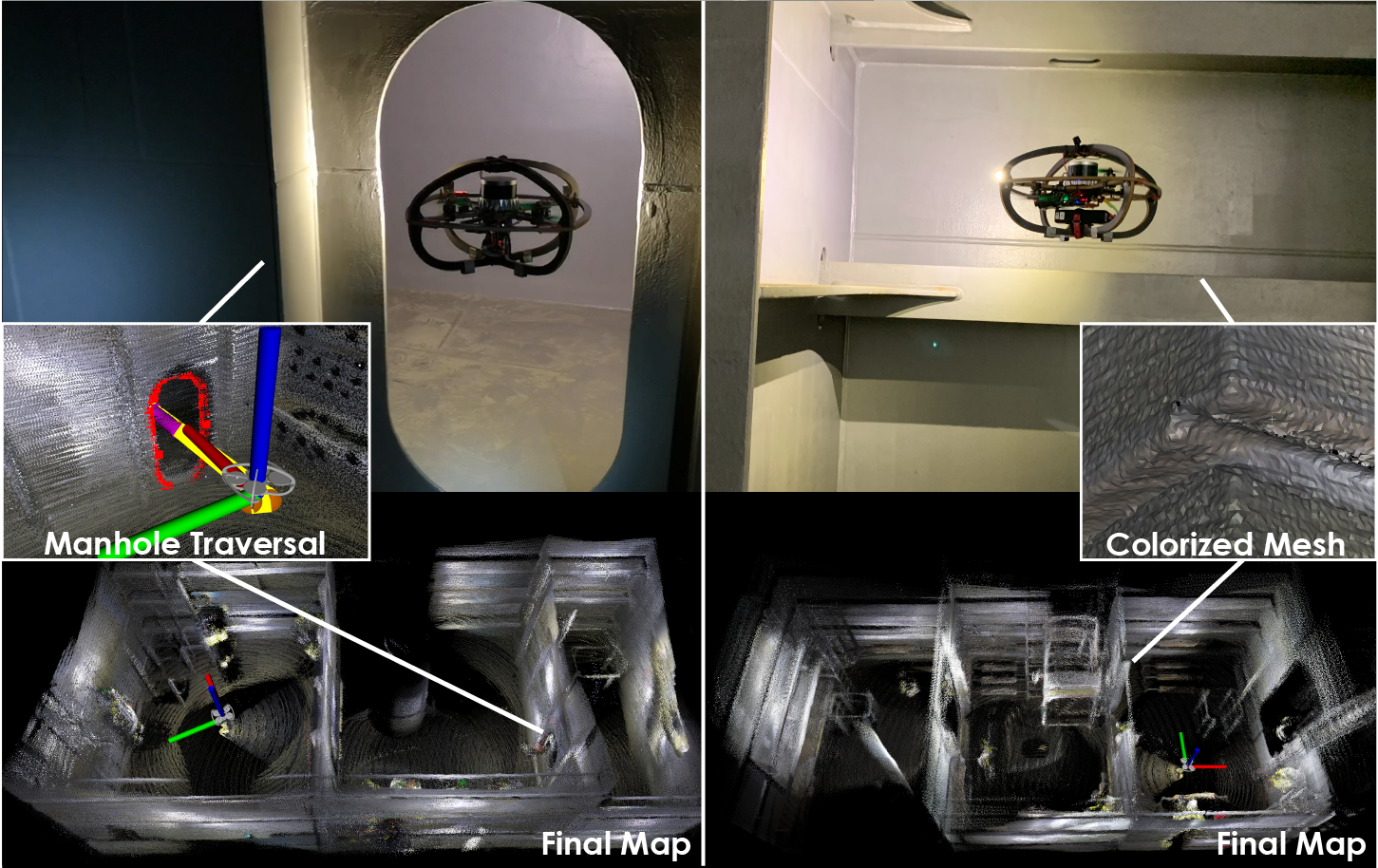}
\caption{Instances of the robot performing autonomous inspection inside the ballast tanks of two different Floating Production Storage and Offloading (FPSO) vessels. Both these missions consisted of the robot mapping and inspecting three compartments of a ballast tank, while the robot had to pass through narrow manholes in the mission on the left.}
\label{fig:intro}
\vspace{-3ex} 
\end{figure}

At the epicenter of the necessary inspection processes is the \ac{gvi}. In simple terms, conventional \ac{gvi} is the process of ``naked eye''-based inspection and detection of damages or anomalies that may pose a risk to the structural integrity and safety of the BWT and thus the vessel as a whole. As \ac{gvi} is often the basis upon which further inspections and maintenance are scheduled, automating this process with robots and enabling the ability for it to take place virtually in any place of the world with little to no human intervention, has the potential to optimize the inspection and maintenance cycles. This in turn will greatly reduce the associated costs, while keeping humans out of harms way~\cite{apostolidis2012modeling}.

Motivated by the above, in this work we present a solution to the problem of automating the general visual inspection of ballast tanks using aerial robots. Importantly, since detailed digital twins and 3D CAD models are generally not available (and will not be available in the foreseeable future) for ships and marine structures due to the longevity of the fleets and other factors like intellectual property handling, we present an approach to enable \ac{gvi} without any prior map of the BWT environments. Instead, the proposed solution enables the combined autonomous geometric exploration and visual inspection that respects certain viewing distance guarantees motivated by the current rulebooks for ship inspection~\cite{IACSrecommendation96,poggi2020recent,dnvRules2022}.

In the remaining paper, Section~\ref{sec:related} presents related work, followed by the problem statement in~\ref{sec:probstat}. The proposed approach is detailed in Section~\ref{sec:approach}, with evaluation studies in Section~\ref{sec:evaluation} and conclusions in Section~\ref{sec:concl}.

\section{RELATED WORK}\label{sec:related}

A niche set of works has approached the problem of robotic inspection of ballast water tanks in ships and other marine structures~\cite{knukkel2023remote,krystosik2021use}. The work in~\cite{carrara2020robotics} presents a set of solutions, including drones and crawlers, on robotic technologies for inspection of ships based on the H2020 European project ROBINS. Recent developments in remote inspection of ship structures are reviewed in~\cite{poggi2020recent}, while the authors in~\cite{caldwell2017hull} discuss hull inspection techniques. The cost-effectiveness of the use of robots for BWT inspection is discussed in~\cite{christensen2011cost}. Reflecting the increasing use of piloted flying robots inside BWT, the authors in~\cite{muhammad2022maritime} overview the associated ecosystem of drone services. Preliminary results on BWT inspection with aerial robots is discussed in~\cite{carrara2020robotics,brogaard2021towards}, while the associated problem of detecting and traversing through narrow manholes is addressed in~\cite{andersen2022depth} and~\cite{manhole2023} using deep learning on dense depth data and a human-engineered detector on LiDAR data respectively. From a different perspective, the authors in~\cite{stensrud2021towards} discuss the utility of hyperspectral imaging onboard drones for BWT inspection. Beyond the use of flying robots for BWT inspection, the works in~\cite{rijnbeek2015rail,tieleman2016rail,christensen2011tank} present approaches on rail-guided robots, while efforts such as those in~\cite{thongpool2015application,andritsos2003rotis} concentrate on the deployment of underwater Remotely Operated Vehicles (ROV) for teleoperated BWT inspection. 

A study of this literature reveals that aerial robotic operations inside BWTs are currently not autonomous but rather typically rely on on-site piloting~\cite{carrara2020robotics}. This is a core limitation. Furthermore, even though piloted robot access can remove some of the safety risks for human personnel, the fact that BWTs often span over multiple compartments implies that RF signals typically used for remote robot control (especially for flying robots) cannot function reliably at the necessary range which in turn means that human personnel still have to be deployed in most areas. As long as trained manned crews have to be onboard the ship involved in the inspection, the dynamics and economics of maritime inspection do not change drastically and will retain their reliance on selected ports around the world and long, expensive, dry-dock operations. A change of paradigm is necessary towards fully autonomous inspection operations. Serving this vision, this work contributes methods and systems to autonomously explore previously unmapped BWTs and simultaneously deliver \ac{gvi} results. To that end, the presented contribution expands upon methods in exploration~\cite{NBVP_ICRA_16,GBPLANNER_JFR_2020,best2022resilient} and inspection path planning~\cite{SIP_AURO_2015,hover2012advanced,galceran2013survey} especially by offering a unifying approach to geometric mapping and visual coverage, while exploiting collision-tolerant miniaturized flying robots~\cite{rmfowl}, and robust onboard localization and mapping~\cite{khattak2020complementary,CERBERUS_WINS_FR2022submission}.

\section{PROBLEM FORMULATION}\label{sec:probstat}

The problem considered in this work is that of combined exploration of a bounded volume $V_E \subset \mathbb{R}^3$ with a depth sensor $\Ys_D$ and simultaneous visual coverage of any desired subset of it $V_C \subset V_E$ with a visual sensor $\Ys_V$ subject to defined viewing distance requirements (motivated by ship inspection requirements~\cite{IACSrecommendation96}). A discrete environment representation is considered through an occupancy map $\mathbb{M}$ consisting of cubical voxels $m\in\mathbb{M}$ with edge length $r_v$. Each voxel $m$ is either unexplored, explored by $\Ys_D$ and free, occupied and explored by $\Ys_D$ but not seen by $\Ys_V$ or occupied and covered by both. If covered by $\Ys_V$ then the distance of observation $\delta$ is also stored. Globally, the problem can be cast as that of a) determining which parts of the initially unexplored volume $V_{une}\overset{init.}{=}V_E$ are free $V_{E,free} \subseteq V_E$ or occupied $V_{E,occ}\subseteq V_E$ (with $\Ys_D$) and b) covering (with $\Ys_V$) all the associated voxels in $V_C$ from a distance $\delta \le \delta_{\max},~\delta_{\max}=\textrm{const}$. The operation is subject to robot motion constraints and the sensors' visibility limitations based on two (distinct) frustum models for $\Ys_D,\Ys_V$ defined by the \ac{fov} $[F_h^D, F_v^D],[F_h^V,F_v^V]$ respectively and maximum ranges $d_{\max}^D,d_{\max}^V$. Hence, certain areas cannot be perceived by $\Ys_D$ or $\Ys_V$ which leads to the following definition of residuals. 

\begin{definition}[Map residuals]\label{def:residualSpace}
 Let $\Xi$ be the simply connected set of collision free configurations $\xi = [x,y,z,\psi]$ and $\bar{\mathcal{V_E}}_m\subseteq \Xi$ the set of all configurations from which the voxel $m$ can be perceived by $\Ys_D$. Then the residual volume is $V_{E,res} = \bigcup_{m\in \mathbb{M}} ( m \vert\ \bar{\mathcal{V_E}}_m = \emptyset )$. Similarly the residual volume $V_{C,res}$ associated with $\Ys_V$ can be defined as $V_{C,res} = \bigcup_{m\in \mathbb{M}} ( m \vert\ \bar{\mathcal{V_C}}_m = \emptyset )$, where $\bar{\mathcal{V_C}}_m\subseteq \Xi$ is likewise the set of all configurations from which $m$ (here a voxel in $V_C$) can be perceived by $\Ys_V$. 
\end{definition}

\begin{problem}[Volumetric Exploration and General Visual Inspection Problem]\label{prob:swapProblem}
 Given a volume $V_E$ and an initial robot configuration $\xi_{init} = [x,y,z,\psi] \subset \Xi$ find a collision-free path $\sigma$ that when traversed by the robot leads to a) identifying $V_{free}$ and $V_{occ}$ based on $\Ys_D$, and b) visually covers all voxels of $V_{occ}$ that are part of $V_C \subset V_E$ from distances $\delta \le \delta_{\max}$. 
\end{problem}

Notably, as in the particular application scenario $V_E$ is considered to incorporate multiple sequential compartments $\{K^i\}$ of ballast tanks, this global problem can be re-cast as that of applying the above process within pairs of volumes $\{V_E^i,V_C^i\},~V_C^i\subset V_E^i$ where $\cup \{V_E^i\} = V_E$ and merging the associated paths $\sigma^i$ by identifying connecting paths traversing through the compartments, potentially via narrow manholes. This characteristic of the problem will be exploited in the proposed approach. Last but not least, the choice of configuration space $\Xi$ is not unique but is opted for given the focus on rotorcraft flying robots in this work. 

\section{PROPOSED APPROACH}\label{sec:approach}
The key steps of the proposed approach for autonomous exploration and \ac{gvi} will be presented in this section.

\subsection{System Architecture} \label{subsec:sys_arch}
\begin{figure}[h!]
\centering
    \includegraphics[width=0.99\columnwidth]{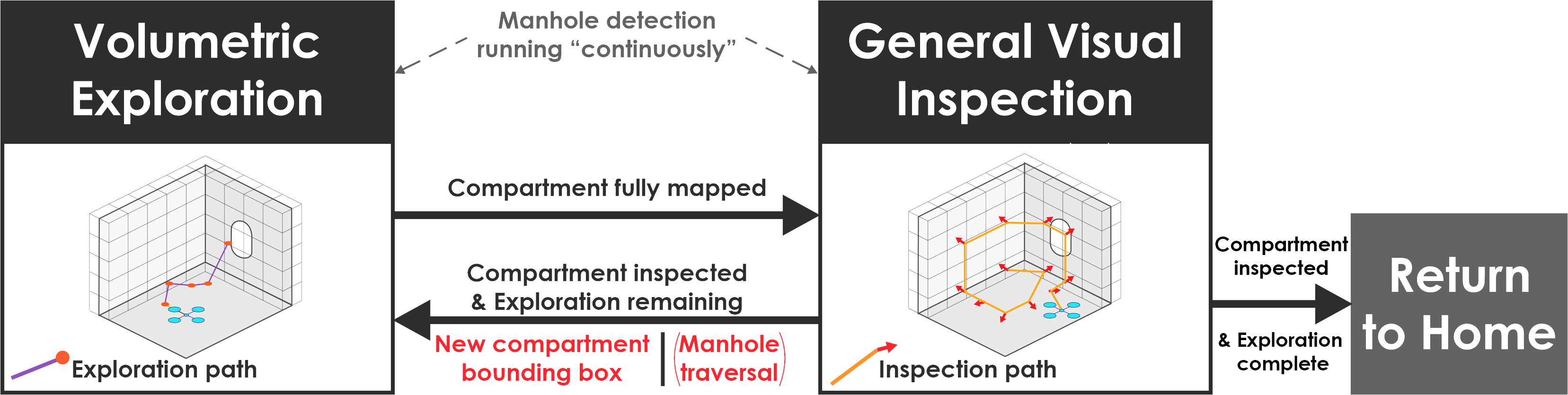} 
\caption{The two operating modes, namely Volumetric Exploration (VE) and General Visual Inspection (GVI), of the system. The robot starts in the VE mode in one compartment, maps that compartment, and then switches to GVI mode to inspect the mapped surface voxels belonging to the compartment with a minimum viewing distance. After completion, it switches back to VE to continue to the next compartment, passing through any detected manholes if required, or returns to the first one if all compartments are inspected.}
\label{fig:flowchart}
\vspace{-3ex} 
\end{figure}

The problem of ballast water tank inspection is formulated as that of exploration and inspection of each compartment, alongside the traversal among such compartments potentially through size-constrained manholes. The robot performs the exploration and general visual inspection of one compartment at a time before moving to the next compartment. The complete task is split into two modes namely, \ac{ve} and General Visual Inspection (GVI). Figure~\ref{fig:flowchart} presents the modes involved in the method and the associated transitions. The planner assumes access to only minimalistic and readily available prior information about two parameters of the tank, namely a) approximate positions of the centers of each compartment, and b) approximate dimensions of the compartments. Such information is known from the $2\textrm{D}$ blueprints that shipowners and operators have readily available, and are also documented from prior inspections. 

The robot starts in one compartment $K^i$ of the ballast tank in the \ac{ve} mode and performs autonomous exploration to map the volume $V^i_E$ corresponding to that compartment fully. Upon completion, it switches to the \ac{gvi} mode in which the planner calculates a single path (as described in Section~\ref{subsec:gvi}) to inspect all surfaces, as represented by the occupied voxels belonging to $V^i_C$, of the mapped compartment using the visual sensor $\Ys_V$ given viewing distance constraints. 
After executing the \ac{gvi} path, the planner switches back to the \ac{ve} mode, to calculate a path leading to the next closest unvisited compartment. To that end, if the ballast tank geometry necessitates, the robot may have to traverse through a manhole that has been previously detected and localized based on our prior work in~\cite{manhole2023}, to reach the next compartment. A specialized planning process is employed to ensure the safe traversal of the particularly narrow manhole openings (often only a few $\textrm{cm}$ wider than the robot itself). The manhole closest to the next compartment to visit is selected, a path is calculated (using the \ac{ve} mode) to reposition the robot in front of that manhole, and upon reaching there the procedure detailed in~\cite{manhole2023} is used to fly through the manhole.
If all compartments are mapped and inspected, the \ac{ve} mode plans a path back to the first compartment as described in Section~\ref{subsec:exploration}.

\subsection{Volumetric Exploration}\label{subsec:exploration}
The \ac{ve} mode utilizes our previous open-sourced work on Graph-based Exploration Path Planning~\cite{GBPLANNER_JFR_2020,GBPLANNER2COHORT_ICRA_2022} called GBPlanner. This method is extensively verified in the field including the winning run of Team CERBERUS in the DARPA Subterranean Challenge Final Event~\cite{CERBERUS_SCIENCE_2022}.
GBPlanner operates in a bifurcated local-global planning architecture. The local exploration stage is responsible for providing efficient exploration paths within a local volume around the current robot location in an iterative fashion. On the other hand, the global respositioning stage is used to reposition the robot to unexplored areas (or other areas of interest) of the map outside the local planner's operational volume. In addition, it is also used to find a safe path for the robot to return to the home location.

The local exploration stage is used in the \ac{ve} mode to map a compartment. For exploring each compartment $K^i$, in each iteration, the planner first builds an undirected graph $\mathbb{G}_L$, along with its vertex and edge sets $\mathbb{V}_L,\mathbb{E}_L$ respectively, within a local volume (in this case slightly larger than the compartment dimensions). Next, the shortest paths $\{\sigma^i_\ell\}$ from the current robot location to all vertices in $\mathbb{G}_L$ are calculated using Dijkstra's algorithm~\cite{dijkstra1959note}. An information gain called the \textbf{VolumeGain}, $\Gamma_{VE}(\nu_k)$, is calculated for each vertex $\nu_k \in \mathbb{V}_L$ and relates to the number of unknown voxels that $\Ys_D$ would perceive if the robot were to be at that vertex. The gain $\Gamma_{VE}(\nu_{j,\ell})$ for each vertex $\nu_{j,\ell} \in \sigma^i_{\ell},j=1...m_{\ell}$ is aggregated along each path $\sigma^i_{\ell}$ to calculate the Exploration Gain $\Lambda_{VE}(\sigma^i_{\ell})$ for that path as: 

\small
\begin{eqnarray}
 \Lambda_{VE}(\sigma^i_{\ell}) = e^{-\zeta \mathcal{Z}(\sigma^i_{\ell},\sigma_{e})} \sum_{j=1}^{m_{\ell}}{\Gamma_{VE}(\nu_{j,\ell}) e^{-\mu \mathcal{D}(\nu_{1,\ell},\nu_{j,\ell})}}
\end{eqnarray}
\normalsize

where $\zeta, \mu>0$ are tunable parameters, $\mathcal{D}(\nu_{1,\ell},\nu_{j,\ell})$ is the cumulative Euclidean distance from vertex $\nu_{j,\ell}$ to the root $\nu_{1,\ell}$ along the path $\sigma^i_{\ell}$, and $\mathcal{Z}(\sigma^i_{\ell},\sigma_{e})$ is a similarity distance metric between $\sigma^i_{\ell}$ as compared to a pseudo straight path $\sigma_{e}$ with the same length along the currently estimated exploration direction. Details are provided in~\cite{GBPLANNER_JFR_2020}. The path $\sigma^i_{VE}$ having the highest Exploration Gain is selected and commanded to the robot. Upon execution, the process is repeated until no vertex in $\mathbb{G}_L$ in an iteration has $\Gamma_{VE}$ greater than a threshold $\Gamma_{VE,\min}$, at which point the planner reports local exploration completion. After completion of the local exploration of a new BWT compartment, the planner switches to the \ac{gvi} mode to facilitate visual inspection. Assuming more compartments are left to be inspected, the method would then act in order to get the robot to the next compartment. For that and other purposes, a global repositioning stage is utilized.

The global repositioning stage of the planner maintains a lightweight graph $\mathbb{G}_G$, built using a subset of the paths $\{\sigma^i_{\ell}\}$ from each local planning iteration, as well as spatially sub-sampled robot poses, within the space explored so far. This graph is used to quickly calculate a path to reposition the robot to areas of the explored map outside the local planner's volume as well as return to the home location. 

This work further uses parts of the global stage of GBPlanner to reposition the robot to the next compartment by calculating a path to the vertex in $\mathbb{G}_G$ closest to its center (or to the entrance of the manhole leading to that compartment).
When all compartments are inspected (or the robot's battery reaches its limit) the graph $\mathbb{G}_G$ is used to calculate a path to the first compartment.

\subsection{General Visual Inspection}\label{subsec:gvi}

After the exploration of a compartment $K^i$ by \ac{ve}, the \ac{gvi} mode of the method is used to perform the visual inspection of all occupied voxels belonging to $V^i_C$ of $K^i$ utilizing the occupancy map built during the \ac{ve} mode. Throughout the operation of the robot, the voxels within this map inside the $\Ys_V$ \ac{fov} and maximum range are annotated as seen voxels. In this mode, the planner calculates a single path which when traversed by the robot will view all unseen voxels belonging to $V^i_C$, while respecting distance constraints.

\begin{algorithm}
\caption{General Visual Inspection}\label{alg:gvi}
\begin{algorithmic}[1]
\State $\mathcal{V}_v \gets \mathbf{generateViewpoints}(\mathbf{B}^i)$
\State $\mathbb{G}_C \gets \mathbf{buildGraph}(\mathcal{V}_v, \xi_{curr})$
\State $\mathbb{G}_C \gets \mathbf{extendGraph}(\mathbb{G}_C,\mathbf{B}^i)$
\State $\mathbf{computeVisualGain}(\mathbb{V}_v)$
\State $\mathbb{L}_s \gets \emptyset,~ \mathbb{L}_r \gets \{\nu_v^k ~|~ \forall \nu_v^k \in \mathbb{V}_v\}$
\While {$\exists ~ \nu^l \in \mathbb{L}_r ~ | ~ \Gamma_v(\nu^l) > \Gamma_{v,min}$}
    \State $\mathbf{sortByVisualGain}(\mathbb{L}_r)$
    \State $\mathbb{L}_s \gets \mathbb{L}_s \cup \mathbb{L}_r[1]$
    \State $\mathbb{L}_r \gets \mathbb{L}_r \setminus \mathbb{L}_r[1]$
    \State $\mathbf{updateVisualGain}(\mathbb{L}_r, \mathbb{L}_s)$
\EndWhile
\State $\mathbb{L}_s \gets \mathbb{L}_s \cup \nu_{curr}$
\State $\mathbb{L}_s \gets \mathbf{orderVertices(\mathbb{L}_s)}$
\State $\sigma^i_{GVI} \gets \mathbf{computeGVIPath(\mathbb{L}_s)}$
\end{algorithmic}
\end{algorithm}

\begin{figure}[h!]
\centering
    \includegraphics[width=0.99\columnwidth]{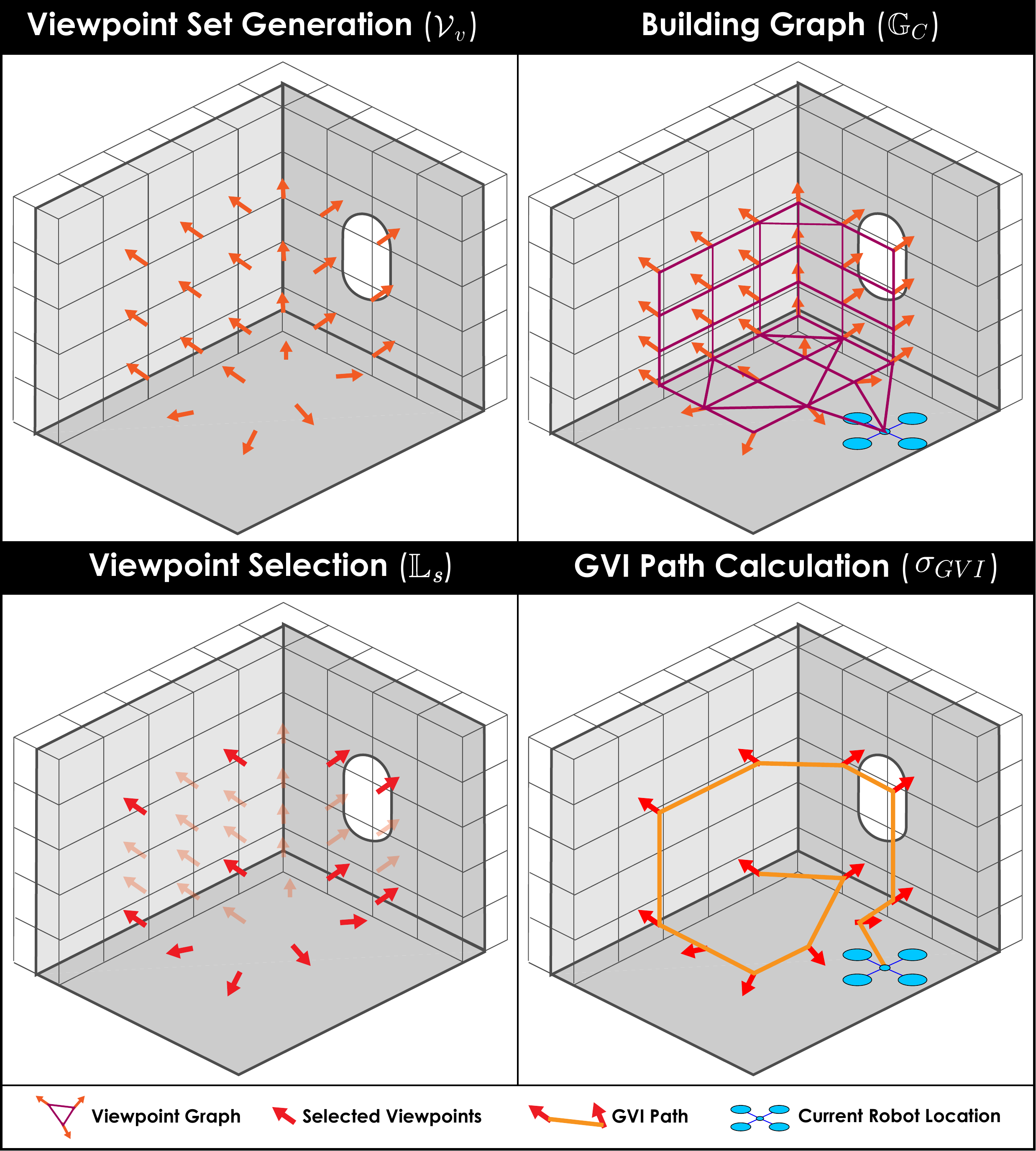}
\caption{Steps involved in calculating the GVI path. A set of viewpoints at a distance in the range $[\delta_{\min}, \delta_{\max}]$ from the occupied voxels are selected in a grid pattern and connected by collision-free straight line edges, as well as sampling additional points if needed, to build a graph. A set of viewpoints covering the entire surface is selected in a greedy manner and their order is calculated by solving the \ac{tsp}.}
\label{fig:gvi_explanation}
\vspace{-1ex} 
\end{figure}

Algorithm~\ref{alg:gvi} describes the steps of \ac{gvi}, while Figure~\ref{fig:gvi_explanation} shows their illustrations. Given the approximate location of the center and the dimensions of the compartment $K^i$, the planner calculates a $3\textrm{D}$ cuboidal bounding box $\mathbf{B}^i$ centered at the compartment center with dimensions slightly larger than those of the compartment. The method first generates a set $\mathcal{V}_v$ of collision-free viewpoints from a uniform 3D grid inside $\mathbf{B}^i$ that have their distance $\partial_s$ to the closest occupied voxel within a range $[\delta_{\min}, \delta_{\max}]$ (line 1), where $\delta_{\min} < \delta_{\max}$ is a constant. It is noted that this distance is readily available through the \ac{esdf} generated by the underlying occupancy map representation, Voxblox~\cite{voxblox}. The orientation of the viewpoints is set to be along the projection of the gradient of the \ac{esdf} at their position, which points to the closest occupied voxel, on the $xOy$ plane of the global coordinate frame. The planner then attempts to connect viewpoints within a range $\rho$ of each other by collision-free straight line edges to build a collision-free graph $\mathbb{G}_C$ (line 2). The current robot state $\xi_{curr}$ is also connected to $\mathbb{G}_C$ in a similar way. As some parts of $\mathbb{G}_C$ might be disconnected from each other, additional points are sampled in $3$D inside $\mathbf{B}^i$ using uniform random sampling and are connected to the vertices in $\mathbb{G}_C$ in order to have no disconnected components (line 3). We refer to the vertices in $\mathbb{G}_C$ corresponding to the viewpoints in $\mathcal{V}_v$ as ``viewpoint vertices'' $\{\nu_v^k\}$ and their set as ``viewpoint vertex set'' $\mathbb{V}_v$. An information gain called \textbf{VisualGain}, $\Gamma_v(\nu_v^k)$, is calculated for each viewpoint vertex $\nu_v^k \in \mathbb{V}_v$ and relates to the number of unseen occupied voxels that would be perceived by $\Ys_V$ if the robot were to be at that vertex (line 4).

To select the set of viewpoints providing complete coverage, we employ a greedy iterative sorting approach. First, all viewpoint vertices are added to a list $\mathbb{L}_r$ (line 5). The list is sorted according to the \textbf{VisualGain} of the vertices (line 7). The vertex with the highest $\Gamma_v$ is selected, removed from $\mathbb{L}_r$, and added to a list $\mathbb{L}_s$ (lines 8-9). The $\Gamma_v$ of the remaining vertices in $\mathbb{L}_r$ is updated to remove overlap with the voxels seen by the vertices in $\mathbb{L}_s$ (line 10). This process is repeated until the \textbf{VisualGain} of all vertices in $\mathbb{L}_r$ is below a threshold $\Gamma_{v,min}$ or $\mathbb{L}_r ~=~ \emptyset$. The vertex $\nu_{curr}$ corresponding to the current robot state, $\xi_{curr}$, is added to $\mathbb{L}_s$ (line 11).
The order in which the vertices in $\mathbb{L}_s$ are to be visited is calculated by solving the Traveling Salesman Problem (TSP) using the Lin-Kernighan-Helsgaun (LKH) heuristic~\cite{helsgaun2000effective}, where the cost of travel between two vertices is defined as the length of the shortest path in $\mathbb{G}_C$ connecting the two vertices (line 12). As the path needs to start at the current robot location but does not need to end at the same location (as will be in the solution of the nominal \ac{tsp}), we set the cost of traveling from any vertex to $\nu_{curr}$ as zero, allowing the path to end at any vertex. The final path to execute, $\sigma^i_{GVI}$, is computed by concatenating the shortest path segments between the consecutive pair of vertices in $\mathbb{L}_s$ in the order given by the solution of the \ac{tsp} starting from $\nu_{curr}$ (line 13).
This path $\sigma^i_{GVI}$ is then commanded to the robot. Upon execution, the planner moves on to the next compartment as described in Section~\ref{subsec:sys_arch}. When all the compartments have been mapped and inspected through the \ac{gvi} mode, the mission is considered complete and the robot automatically returns to the first compartment.

\section{EVALUATION STUDIES}\label{sec:evaluation}
To evaluate the performance of the method, we present a simulation study along with three experiments in real ships.

\begin{figure}[h!]
\centering
    \includegraphics[width=0.99\columnwidth]{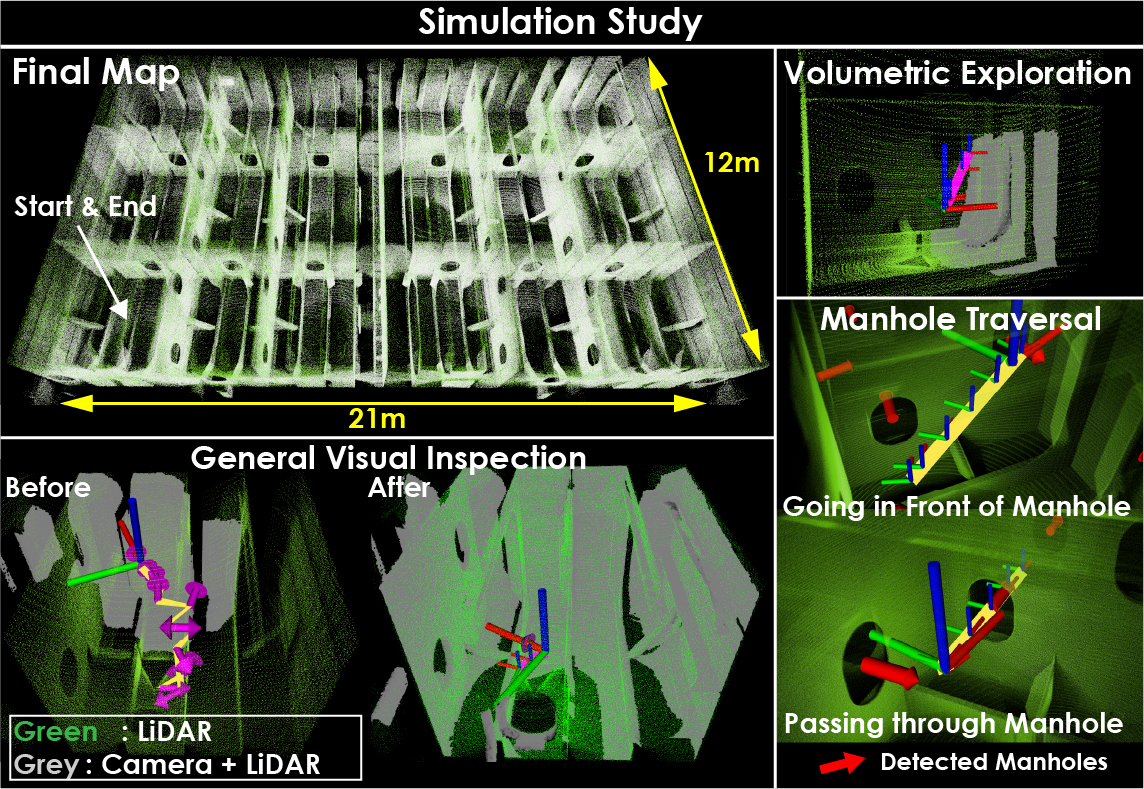}
\caption{Results of the simulation study conducted in a $3\textrm{D}$ CAD model of a ship inside a part of a double bottom ballast tank. The robot explored and inspected $18$ compartments in total, passing through manholes of dimensions $\textrm{height} \times \textrm{width} = \SI{0.8}{\meter} \times \SI{0.6}{\meter}$, resulting in a $20$ minute-long mission.}
\label{fig:sim}
\vspace{-3ex} 
\end{figure}

\begin{figure*}[h!]
\centering
    \includegraphics[width=0.99\textwidth]{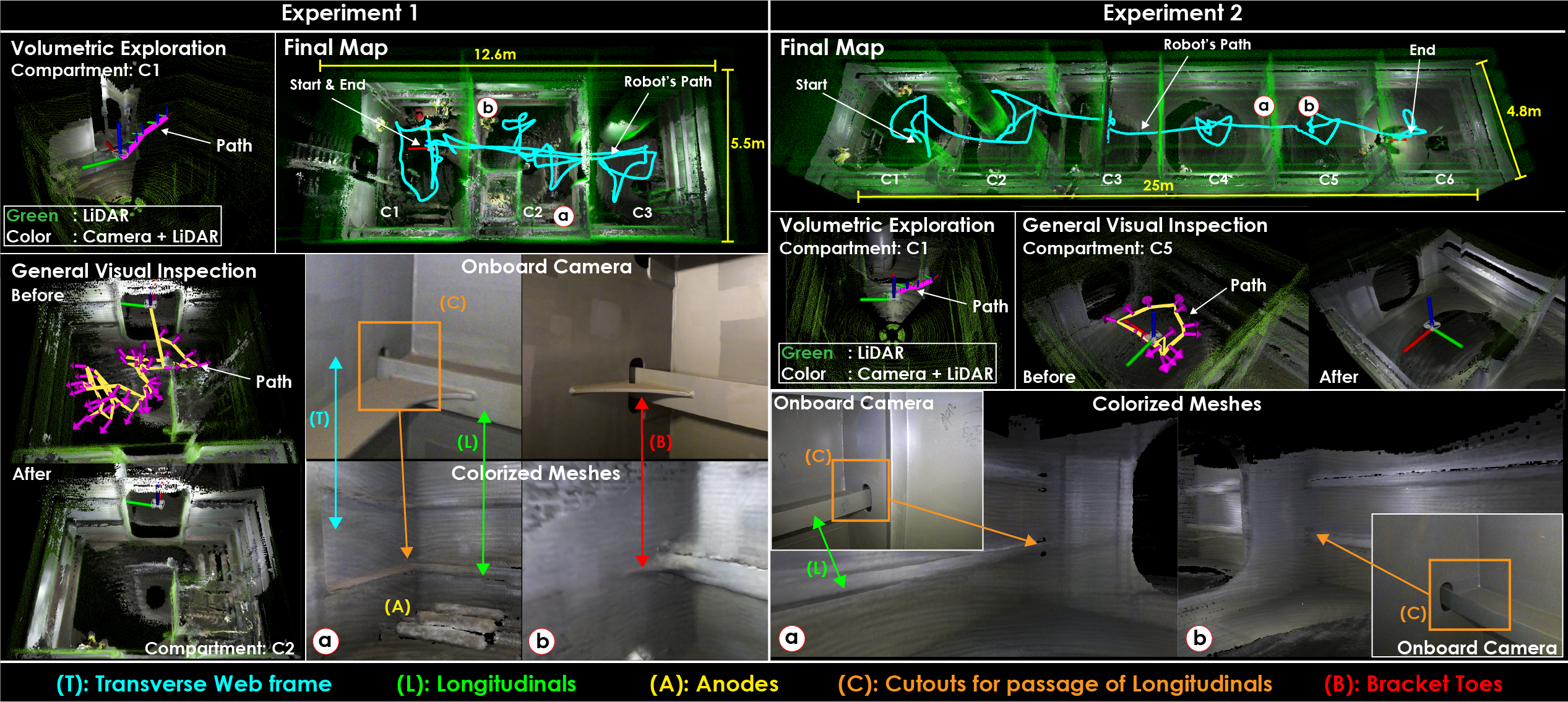}
\caption{Results of experiments 1 and 2 conducted in Vessel1 and Vessel2 respectively. The two experiments required the robot to explore and inspect $3$ and $6$ compartments respectively, that were connected by large openings. The figure shows the final map and path traversed by the robot, and indicative instances of VE and GVI. The visual coverage before and after the GVI step in one compartment is also shown through point cloud colorization (green point color reserved for areas not observed by the camera due to imposed robot height constraints and inaccessible areas). Furthermore, instances of the robot viewing inspection important structures can be seen, with the corresponding the onboard camera images and parts of reconstructed meshes colorized by camera images.}
\label{fig:large_openings_exp}
\vspace{-3ex} 
\end{figure*}

\subsection{Simulation Studies} 
To validate and fine-tune the method, we conducted a large-scale simulation study demonstrating the capability of the autonomous exploration and inspection system. The study took place in a $3\textrm{D}$ CAD model of a ship inside a part of a double-bottom ballast tank. The selected area consisted of $18$ compartments in a $6 \times 3$ grid connected by multiple manholes as can be seen in Figure~\ref{fig:sim}. The dimensions of the compartments were $\textrm{l} \times \textrm{w} \times \textrm{h} = \SI{4.0}{\meter} \times \SI{3.5}{\meter} \times \SI{3.0}{\meter}$ and those of the manholes were $\textrm{height} \times \textrm{width} = \SI{0.8}{\meter} \times \SI{0.6}{\meter}$. It is noted that a single mission of such scale ($18$ compartments) is not required in practice as it can be split into multiple missions as per the robot's battery limitations and with the system entering from different access hatches. However, it was opted to run such a large-scale simulation to thoroughly evaluate the performance of the proposed inspection solution.
We used the RotorS~\cite{furrer2016rotors} simulator with a model of the RMF-Owl~\cite{rmfowl} robot having the same sensors as the real platform (check Section~\ref{subsubsec:rmf} for further details).
The robot started in one of the compartments, performed exploration and inspection of each compartment, repositioning to unvisited compartments by traversing through the manholes, and finally returned to the first compartment after inspecting all compartments at a nominal speed of \SI{1.0}{\meter/s}. Figure~\ref{fig:sim} provides details of the mission including instances of \ac{ve} and \ac{gvi}, manhole traversal, and the final map of the tank with the surfaces inspected by the robot annotated in a different color. 


\subsection{Field Experiments}
To demonstrate the method's applicability in the real world, we present three experiments conducted in the ballast tanks of two \ac{fpso} vessels, hereby called ``Vessel 1'' and ``Vessel 2'', including navigation through manholes. Access to the two vessels was provided by Equinor ASA and Altera Infrastructure Production AS respectively. The experiments consisted of autonomous exploration and inspection of multiple compartments of BWTs and were conducted using a modified version of the RMF-Owl robot~\cite{rmfowl} navigating at a nominal speed of \SI{0.85}{\meter/s}. The description of the robot and the experiments is given in the remaining section.

\begin{figure}[h!]
\centering
    \includegraphics[width=0.99\columnwidth]{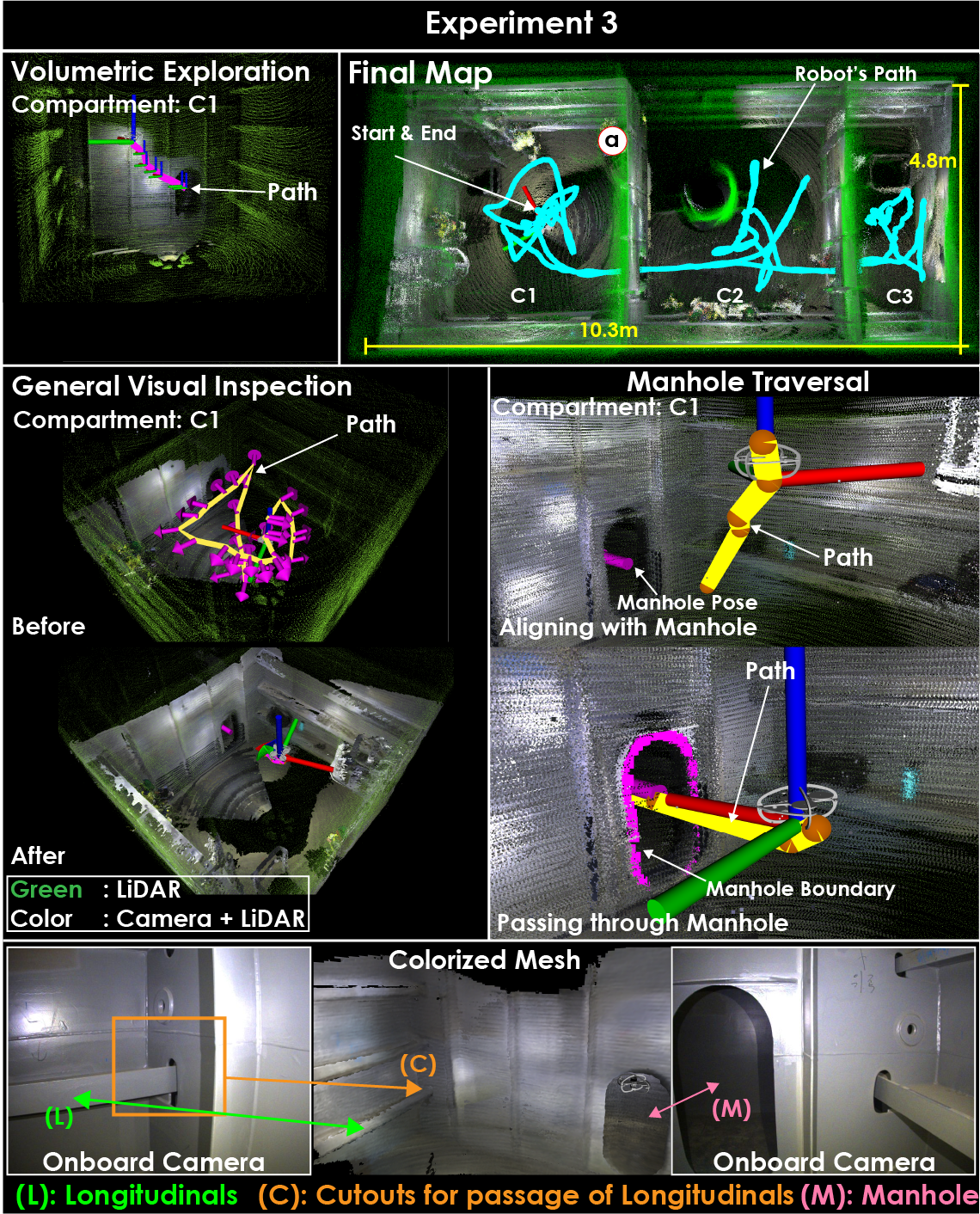}
\caption{Results of Experiment 3 at Vessel 2. The robot explored and inspected $3$ compartments connected by narrow manholes ($\textrm{height} \times \textrm{width} = \SI{1.3}{\meter} \times \SI{0.6}{\meter}$). The robot performed explicit detection and localization of the manholes to calculate a path through them as shown in the figure. The final path and map, instances of VE, GVI, and manhole traversal, as well as the robot viewing inspection-important areas, can be seen.}
\label{fig:manhole_exp}
\vspace{-4ex} 
\end{figure}

\begin{figure*}[h!]
\centering
    \includegraphics[width=0.99\textwidth]{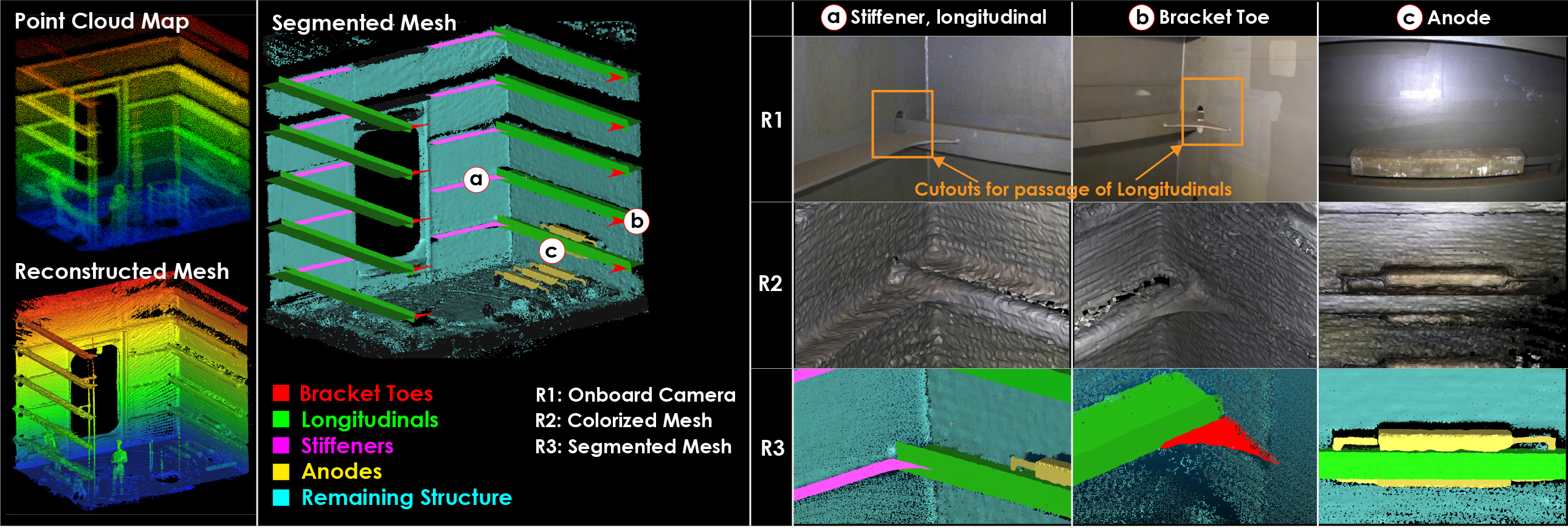}
\caption{Post processed inspection result of one compartment from Experiment 1. The result shows the complete point cloud and mesh, semantically segmented mesh, as well as certain semantic structures with the corresponding onboard camera images. The inspection important structures (semantics) selected for this result were ``Longitudinals'', ``Brackets'' between the longitudinals and transverse web frames, ``Stiffeners'', and the ``Anodes''. This shows how such a mission can be processed to generate digital models of inspection important areas which can be associated with robot's path and onboard camera images to perform detailed inspection either by a human or possibly automated algorithms to detect corrosion and other risk factors.}
\label{fig:inspection_result}
\end{figure*}

\subsubsection{System Overview}\label{subsubsec:rmf}
The robot utilized for the conducted experiments is a modified version of the RMF-Owl system (Figure~\ref{fig:intro}). Protected by a custom-designed frame made of carbon-foam sandwich material, RMF-Owl is designed for accessing confined areas, with robust and collision-tolerant flight capabilities. It weighs $\SI{1.45}{\kilogram}$ and has dimensions $\textrm{length}\times\textrm{width}\times\textrm{height} ~=~ \SI{0.38}{\meter} \times \SI{0.38}{\meter} \times \SI{0.24}{\meter}$. It utilizes a 4s 5000 mAh LiPo battery and has an endurance of $10$ min. The sensing payload of RMF-Owl is tailored for the combined objective of autonomous volumetric exploration and general visual inspection. Specifically, it integrates a sensing suite consisting of an Ouster OS0-64 LiDAR with \ac{fov} $[F^D_h, F^D_v]=[360,90]^\circ$ and a Flir Blackfly S color camera with $[F^V_h, F^V_v]=[85,64]^\circ$, as well as a VectorNav VN-100 IMU interfaced with a Khadas VIM4 Single Board Computer (SBC) incorporating $\times 4$ 2.2Ghz Cortex-A73 cores, paired with $\times4$ 2.0Ghz Cortex-A53 cores implementing an Amlogic A311D2 big-little architecture. The Khadas SBC interfaces the PixRacer autopilot to which it provides navigation commands. The system is further enhanced with the DJI O3 camera enabling the onboard recording of 4K video up to $120$ FPS. 

\subsubsection{Experiment 1}
The first experiment, conducted in Vessel 1, involved $3$ compartments of a ballast tank connected by large openings. The dimensions of each compartment were $\textrm{l} \times \textrm{w} \times \textrm{h} = \SI{5.5}{\meter} \times \SI{4.2}{\meter} \times \SI{5.0}{\meter}$. The robot started in the first compartment, performed exploration and inspection of each compartment, and upon completion of all three of them, it returned to the starting location in the first compartment. Table~\ref{tab:exp_stats} shows the quantitative results and parameters related to the experiments including the total volume mapped $V_{E,tot}$, total surface inspected $S_{C,tot}$, percentage of the mapped surface seen by the camera $\lambda_C$, and the surface density of the final map $\varrho$ in terms of LiDAR $\textrm{points/}\textrm{m}^2$. Figure~\ref{fig:large_openings_exp} - Experiment 1, shows the details of the experiment. Instances of \ac{ve} and \ac{gvi}, the final map annotated with parts inspected by the robot, as well as offline mesh reconstructions of parts of the environment and the corresponding areas viewed by the onboard camera, can be seen. The subfigure titled ``Colorized Meshes'' shows instances of three structures that are important for inspection as prescribed by the \ac{iacs}~\cite{IACSrecommendation96}. The structures are namely, ``Longitudinals'', ``Brackets'' between the longitudinals and transverse web frames, and the ``Anodes''. The first two are highly susceptible to cracks and corrosion, and the third is important to inspect for depletion. Furthermore, the cutout made in the transversal web frame for the passage of longitudinals is prone to cracking and is also highlighted in the figure. Images from the onboard camera viewing these structures within the commanded viewing distance are also shown in Figure~\ref{fig:large_openings_exp}. 

\subsubsection{Experiment 2}
The second experiment was conducted in Vessel 2. It involved $6$ compartments of dimensions $\textrm{l} \times \textrm{w} \times \textrm{h} = \SI{4.8}{\meter} \times \SI{4.16}{\meter} \times \SI{3.0}{\meter}$, connected by large openings similar to Experiment 1. The robot explored and inspected all six compartments but was commanded not to return to the first compartment due to battery limitations (the experiment was initiated with a not fully charged battery). The details of the experiment are shown in Figure~\ref{fig:large_openings_exp} - Experiment 2 and the statistics in Table~\ref{tab:exp_stats}.

\subsubsection{Experiment 3}
Finally, the third experiment also took place in Vessel 2 but in a section of the ballast tank having $3$ compartments connected by manholes of dimensions $\textrm{height} \times \textrm{width} = \SI{1.3}{\meter} \times \SI{0.6}{\meter}$. The dimensions of the first two compartments were $\textrm{l} \times \textrm{w} \times \textrm{h} = \SI{4.8}{\meter} \times \SI{4.2}{\meter} \times \SI{5.0}{\meter}$ whereas the third compartment was smaller having dimensions $\textrm{l} \times \textrm{w} \times \textrm{h} = \SI{4.8}{\meter} \times \SI{2.0}{\meter} \times \SI{3.0}{\meter}$. The robot performed exploration and inspection of all three compartments, passing through the manholes, and returned to its starting location upon completion of the mapping mission. Figure~\ref{fig:manhole_exp} shows the final map, \ac{ve} and \ac{gvi} paths, and colorized mesh of a part of a compartment along with onboard camera images. 

The data collected from one of the missions was further processed offline to segment out inspection-important structures and create mesh reconstructions of those and assess the complete data package involving geometric reconstructions and visual data. Such processed data can be used for analyzing the inspection result, as well as planning future inspection missions exploiting prior maps. 
For this work we focused on the following inspection-important structures ``Longitudinals'', ``Brackets'' between the longitudinals and transverse web frames, ``Stiffeners'', and the ``Anodes''. The structures were manually segmented from the aggregated point cloud of the BWTs and were filtered using the statistical outlier removal method to reduce noise. As most of these structures are planar or can be split into a set of planar surfaces, such planes were extracted using RANSAC and the points corresponding to each plane were projected on it to create a clean point cloud. The cleaned point clouds were then used to create the mesh reconstructions. Then given the robot path and the collected visual camera data, the inspected geometries can be related to the associated camera frames enabling detailed inspection by a human or possibly automated algorithms to detect corrosion and other risk factors. Figure~\ref{fig:inspection_result} shows the end result of this process for one compartment from Experiment 1. In the left half, the aggregated point cloud map, mesh reconstruction of the entire point cloud, and the segmented mesh can be seen. The right half of the figure shows instances of the robot inspecting these structures. The onboard camera image, colorized mesh based on camera images, and segmented mesh is shown. 
This allows a potential tank inspector to easily identify the inspection important structures, visualize their geometric reconstructions, and view the associated camera image captured by the robot.

\begin{table}[]
\centering
\vspace{-1ex}
\caption{Quantitative results for the experiments.}
\begin{tabular}{|l|l|l|l|}
\hline
\textbf{Parameter}                                  & \textbf{Experiment1}   & \textbf{Experiment2}   & \textbf{Experiment3}   \\ \hline
$V_{E,tot}$ ($\textrm{m}^3$)                        & $343.2$                & $429.5$                & $207.5$                \\ \hline
$S_{C,tot}$ ($\textrm{m}^2$)                        & $288.89$               & $477.19$               & $299.47$                \\ \hline
$\lambda_C$ ($\textrm{\%}$)                         & $87.95$                & $95.99$                & $92.48$                \\ \hline
$\varrho$ ($\textrm{points/}\textrm{m}^2$)          & $102,972$              & $115,883$              & $152,158$                \\ \hline
$\delta_{\max}$ ($\textrm{m}$)                      & $1.25$                 & $1.5$                  & $1.25$                \\ \hline
\end{tabular}
\label{tab:exp_stats}
\vspace{-3ex}
\end{table}





\section{CONCLUSIONS}\label{sec:concl}
In this paper, a method for combined volumetric exploration and general visual inspection of ballast water tanks is presented. Tailored to the geometry and topology of the ballast tanks, the method is able to perform inspection from desired viewing distances from surfaces with no prior map. Both simulation and field experiments are presented to verify the proposed solution. The paper further shows that the data from these missions can be used to generate inspection-relevant results such as semantically segmented meshes of the tanks, and associate the geometric structures with onboard camera images for detailed analysis.


\addtolength{\textheight}{-4cm}   




\bibliographystyle{IEEEtran}
\bibliography{./BWTGVI_ICAR_2023}

\end{document}